%% file: main.tex
\documentclass[10pt,twocolumn,letterpaper]{article}

\usepackage[pagenumbers]{cvpr}

\definecolor{cvprblue}{rgb}{0.21,0.49,0.74}
\usepackage[pagebackref,breaklinks,colorlinks,allcolors=cvprblue]{hyperref}
\usepackage{multirow}
\usepackage{colortbl}
\usepackage{booktabs}
\usepackage{float}
\usepackage{tocloft}
\usepackage{fontawesome}
\usepackage{bbding}
\usepackage{url}

\title{YOLO-UniOW: Efficient Universal Open-World Object Detection}

\author{
Lihao Liu$^{1,\*}$\footnotemark[1],~
Juexiao Feng$^{1,\*}$\footnotemark[1],~
Hui Chen$^{1}$,~
Ao Wang$^{1}$,~
Lin Song$^{2}$,~
Jungong Han$^{1}$,~
Guiguang Ding$^{1, \text{\Envelope}}$ \\
$^1$~Tsinghua University~
$^2$~Tencent ARC Lab
}

\begin{document}
\maketitle
{
\renewcommand{\thefootnote}{\fnsymbol{footnote}}
\footnotetext[1]{Equal contribution. \text{\Envelope} Corresponding author.}
}
\input{sec/0_abstract}    
\input{sec/1_intro}
\input{sec/2_Related_Work}
\input{sec/3_Method}
\input{sec/4_Experiments}

\input{sec/5_Conclusion}
{
    \small
    \bibliographystyle{ieeenat_fullname}
    \bibliography{main}
}

\end{document}

%% file: sec/0_abstract.tex
\begin{abstract}
Traditional object detection models are constrained by the limitations of closed-set datasets, detecting only categories encountered during training. While multimodal models have extended category recognition by aligning text and image modalities, they introduce significant inference overhead due to cross-modality fusion and still remain restricted by predefined vocabulary, leaving them ineffective at handling unknown objects in open-world scenarios.
In this work, we introduce Universal Open-World Object Detection (Uni-OWD), a new paradigm that unifies open-vocabulary and open-world object detection tasks. To address the challenges of this setting, we propose YOLO-UniOW, a novel model that advances the boundaries of efficiency, versatility, and performance. YOLO-UniOW incorporates Adaptive Decision Learning to replace computationally expensive cross-modality fusion with lightweight alignment in the CLIP latent space, achieving efficient detection without compromising generalization. Additionally, we design a Wildcard Learning strategy that detects out-of-distribution objects as ``unknown" while enabling dynamic vocabulary expansion without the need for incremental learning. This design empowers YOLO-UniOW to seamlessly adapt to new categories in open-world environments.
Extensive experiments validate the superiority of YOLO-UniOW, achieving achieving 34.6 $\text{AP}$ and 30.0 $\text{AP}_r$ on LVIS with an inference speed of 69.6 FPS. The model also sets benchmarks on M-OWODB, S-OWODB, and nuScenes datasets, showcasing its unmatched performance in open-world object detection. Code and models are available at \url{https://github.com/THU-MIG/YOLO-UniOW}.

\end{abstract}

%% file: sec/1_intro.tex
\section{Introduction}
\label{sec:intro}

Object detection has long been one of the most fundamental and widely applied techniques in the field of computer vision, with extensive applications in security~\cite{shen2024x}, autonomous driving~\cite{yang2023gpro3d}, and medical imaging~\cite{guo2022deep}. Many remarkable works have achieved breakthroughs for object detection, such as Faster R-CNN~\cite{faster_rcnn}, SSD~\cite{SSD}, RetinaNet~\cite{Retinanet}, etc. 

\begin{figure}
    \centering
    \includegraphics[width=\linewidth]{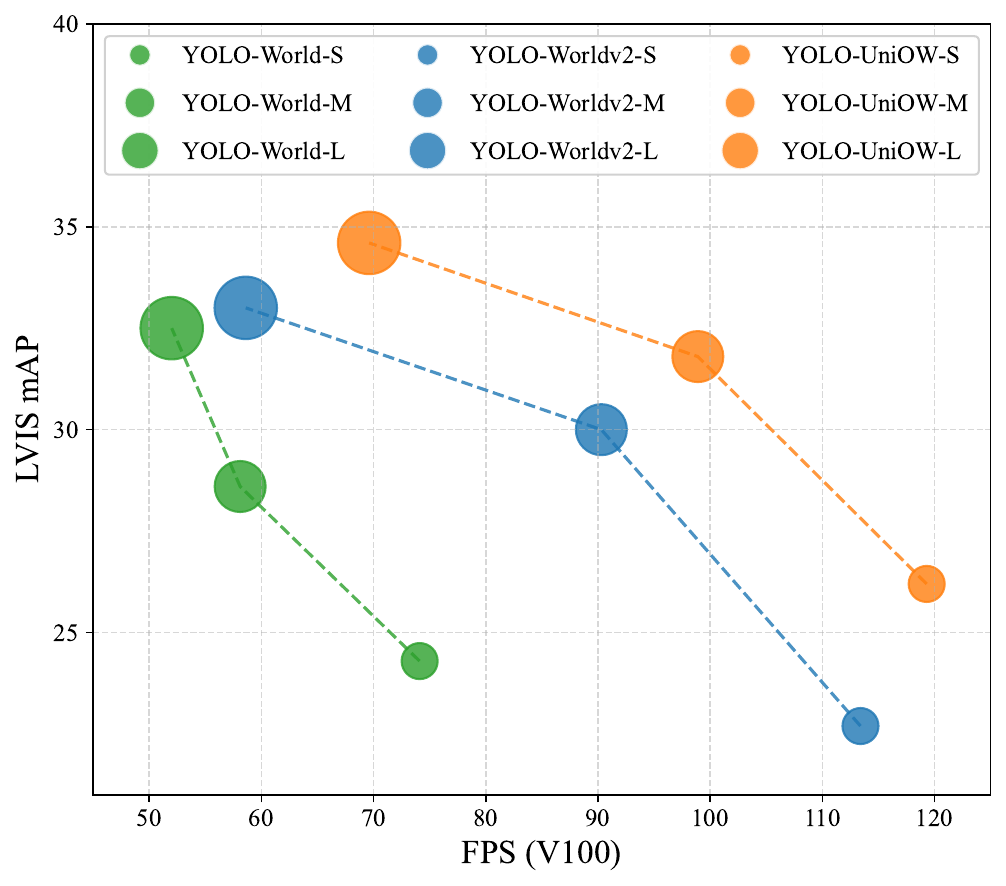}
    \caption{\textbf{Speed-Accuracy Trade-off Curve}. Comparison of YOLO-UniOW and recent methods in speed and accuracy on the LVIS minival dataset. Inference speed is measured on a single NVIDIA V100 GPU without TensorRT. Circle size indicates model size.}
    \label{fig:FPS-LVIS}
\end{figure}

\begin{figure*}
    \centering
    \includegraphics[width=\linewidth]{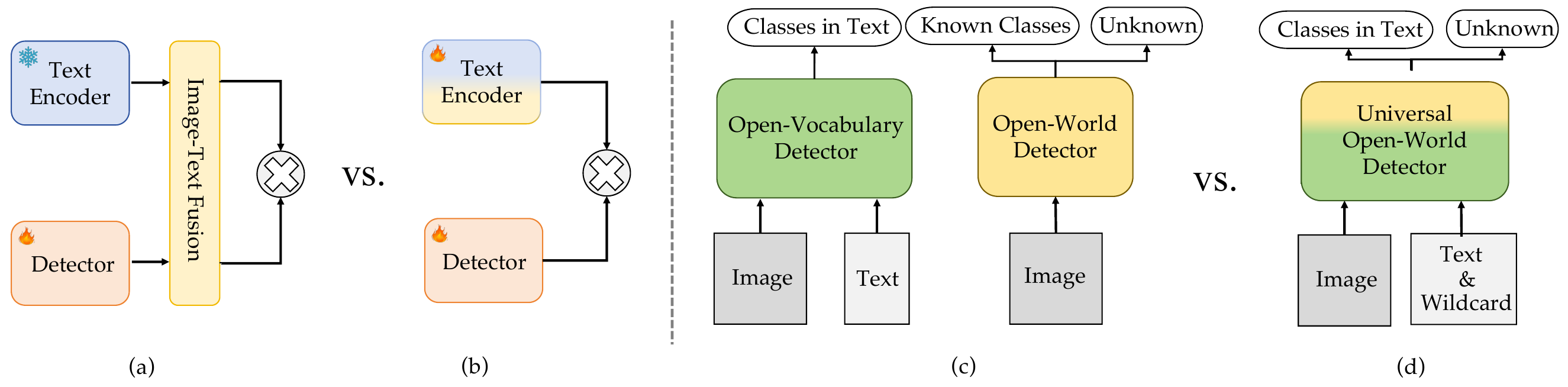}
    \caption{\textbf{Comparisons of Detection Framework.} (a) Open-vocabulary detector with cross-modality fusion. (b) Our efficient open-vocabulary detector with Adaptive Decision Learning. (c) Open-world and open-vocabulary detectors. (d) Our Uni-OWD detector for both open-vocabulary and open-world tasks.}
    \label{fig:framework}
\end{figure*}

In recent years, the YOLO (You Only Look Once)~\cite{yolov3,yolov4,yolov8,yolov10} series of models has gained widespread attention for its outstanding detection performance and real-time efficiency. The recent YOLOv10~\cite{yolov10} establishes a new standard for object detection by employing a consistent dual assignment strategy, achieving efficient NMS-free training and inference. 

However, traditional YOLO-based object detection models are often confined to a closed set definition, where objects of interest belong to a predefined set of categories. 

In practical open-world scenarios, when encountering \textit{unknown} categories that have not been encountered in the training datasets, these objects are often misclassified as background. This inability of models to recognize novel objects can also negatively impact the accuracy of known categories, limiting their robust application in real-world scenarios.

Thanks to the development of vision-language models, such as ~\cite{clip,align,evaclip,chen2020imram}, combining their open-vocabulary capabilities with the efficient object detection of YOLO presents an appealing and promising approach for real-time open-world object detection. YOLO-World~\cite{yoloworld} is a pioneering attempt, where YOLOv8 \cite{yolov8} is used as the object detector, and CLIP’s text encoder is integrated as an open-vocabulary classifier for region proposals (i.e., anchors in YOLOv8). The decision boundary for object recognition is derived from representations of class names generated by CLIP's text encoder. Additionally, a vision-language path aggregation network (RepVL-PAN) using re-parameterization ~\cite{ding2021repvgg,wang2024repvit} is introduced to comprehensively aggregate text and image features for better cross-modality fusion.

Although YOLO-World is effective for open-vocabulary object detection (OVD), it still relies on a predefined vocabulary of class names, which must include all categories that are expected to be detected. This reliance significantly limits its ability to dynamically adapt to newly emerging categories, as determining unseen class names in advance is inherently challenging, preventing it being truly open-world. Moreover, the inclusion of RepVL-PAN introduces additional computational costs, especially with large vocabulary sizes, making it less efficient for real-world applications.

In this work, we first advocate a new setting of \textbf{Universal Open-World Object Detection (Uni-OWD)}, in which we encourage realizing open-world object detection (OWOD) and open-vocabulary object detection (OVD) with one unified model. Specifically, it emphasizes that the model can not only recognize categories unseen during training but also effectively classify unknown objects as ``unknown''. Additionally, we call for a efficient solution following YOLO-World to meet the efficiency requirement in real-world applications. To achieve these, we propose a \textbf{YOLO-UniOW} model to achieve effective universal open-world detection but also enjoying greater efficiency.

Our YOLO-UniOW emphasizes several insights for efficient Uni-OWD. (1) \textbf{Efficiency}. Except using recent YOLOv10~\cite{yolov10} for the more efficient object detector, we introduce a novel adaptive decision learning strategy, dubbed AdaDL, to wipe out the expensive cross-modality vision-language aggregation in RepVL-PAN, as illustrated in \cref{fig:framework} (b). The goal of AdaDL is to adaptively capture task-related decision representations for object detection without sacrificing the generalization ability of CLIP. Therefore, we can well align the image feature and class feature directly in the latent CLIP space with no any heavy cross-modality fusion operations, achieving efficient and outstanding detection performance (see \cref{fig:FPS-LVIS}). (2) \textbf{Versatility}. The challenge of open-world object detection (OWOD) lies in differentiating all unseen objects with only one ``unknown'' category \textit{without any supervision about unknown objects}. To solve this issue, we design a wildcard learning method that use a wildcard embedding to unlock generic power of open-vocabulary model. This wildcard embedding is optimized through a simple self-supervised learning, which seamlessly adapts to dynamic real-world scenarios. As  shown in \cref{fig:framework} (d), our YOLO-UniOW can not only benefit from the dynamic expansion of the known category set like YOLO-World, \ie, open-vocabulary detection, but also can highlight any out-of-distribution objects with ``unknown'' category for open-world detection.  (3) \textbf{High performance}. We evaluated our zero-shot open-vocabulary capability in LVIS~\cite{livs}, and the open-world approach in benchmarks such as M-OWODB~\cite{scheirer2012toward}, S-OWODB~\cite{s-owodb}, and nuScenes~\cite{nuScenes}. Experimental results show that our method can significantly outperform existing state-of-the-art methods for efficient OVD, achieving 34.6 $\text{AP}$, 30.0 $\text{AP}_r$ on the LVIS dataset with the speed of 69.6 FPS. Besides, YOLO-UniOW can also perform well in both zero-shot and task-incremental learning for open-world evaluation. These well demonstrate the effectiveness of the proposed YOLO-UniOW. 

 The contributions of this work are as follows:

\begin{itemize}
    \item We advocate a new setting of Universal Open-World Object Detection, dubbed Uni-OWD to solve the challenges of dynamic object categories and unknown target recognition with one unified model. We provide an efficient solution based on YOLO detector, ending up with our YOLO-UniOW.
    
    \item We design a novel adaptive decision learning (AdaDL) strategy to adapt the representation of decision boundary into the task of Uni-OWD without sacrificing the generalization ability of CLIP. Thanks to AdaDL, we can leave out the heave computation of cross-modality fusion operation used in previous works.
    
    \item We introduce wildcard learning to detect unknown objects, enabling iterative vocabulary expansion and seamless adaptation to dynamic real-world scenarios. This strategy eliminates the reliance on incremental learning strategies.

    \item Extensive experiment across benchmark for both open-vocabulary object detection and open-world object detection show that YOLO-UniOW can significantly outperform existing methods, well demonstrating its versatility and superiority.
\end{itemize}

%% file: sec/2_Related_Work.tex
\section{Related Work}
\label{sec:related}

\begin{figure*}[t]
    \centering
    \includegraphics[width=\linewidth]{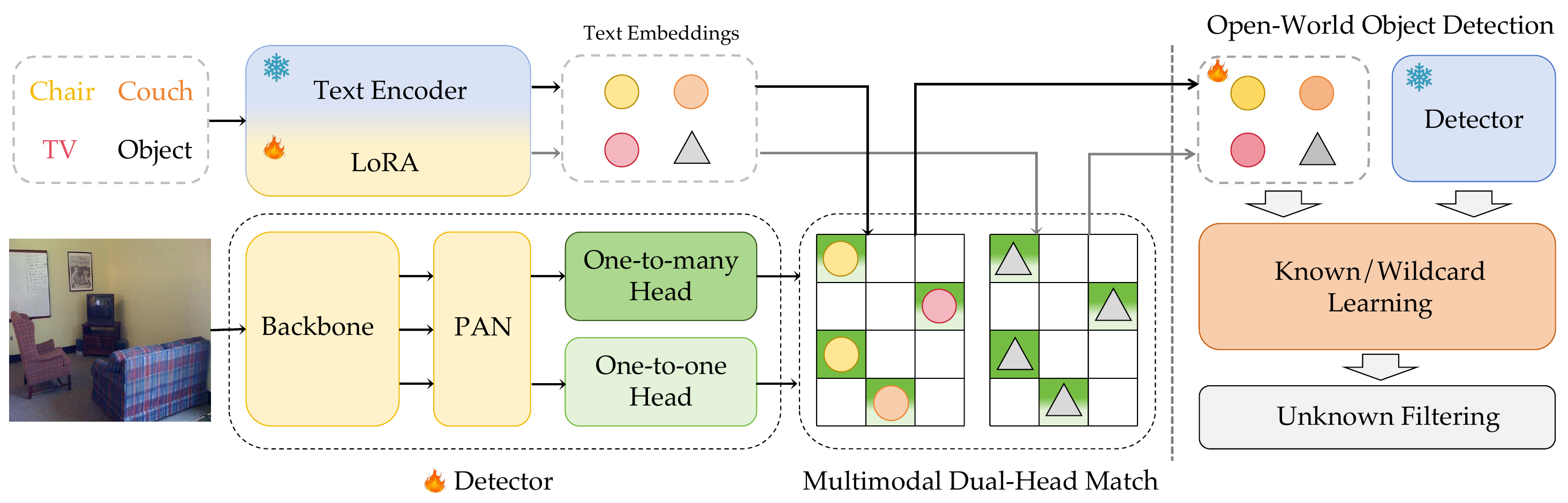}
    \caption{\textbf{Our Proposed Efficient Universal Open-World Object Detection Pipeline.} Open-Vocabulary 
    Pretraining (left): Using a \textit{Multimodal Dual-Head Match} for efficient end-to-end object detection, \textit{AdaDL} in text encoder for adaptive decision boundary learning. Open-World Fine-tuning (right): Utilizing calibrated text embeddings and the detector to adaptively detect both known and unknown objects with the assistance of the wildcard. A filtering strategy is employed to remove duplicate unknown predictions, ensuring efficient and effective open-world object detection.}
    \label{fig:overall}
\end{figure*}

\subsection{Open-Vocabulary Object Detection}
Open-Vocabulary Object Detection (OVD) has emerged as a prominent research direction in computer vision in recent years. Unlike traditional object detection, OVD enables the detection dynamically expand categories without relying heavily on the fixed set of categories defined in the training dataset. Several works have explored leveraging Vision-Language Models (VLMs) for enhancing object detection. For instance, ~\cite{glip,glipv2,detclip,detclipv2,groundingdino,grounding15,RegionCLIP,codet,yoloworld} utilize large-scale, easily accessible text-image pairs for pretraining, resulting in more robust and generalizable detectors, which are subsequently fine-tuned on specific target datasets. In parallel, ~\cite{vild,hierdk,lpovod,clipself} focus on distilling the alignment of visual-text knowledge from VLMs into object detection, emphasizing the design of distillation losses and the generation of object proposals. Additionally, ~\cite{detpro,promptdet,cora} investigate various prompt modeling techniques to more effectively transfer VLM knowledge to the detector, enhancing its performance in open-vocabulary and unseen category tasks.

\subsection{Open-World Object Detection}
Open-World Object Detection (OWOD) is an emerging direction in object detection, aiming to address the challenge of dynamic category detection. The goal is to enable detection models to identify known categories while recognizing unknown categories, and to incrementally adapt to new categories over time. Through methods such as manual annotation or active learning ~\cite{roy2018deep, yuan2021multiple,boxlevel}, unknown categories can be progressively converted into known categories, facilitating continuous learning and adaptation.

The concept of OWOD was first introduced by Joseph et al.~\cite{joseph2021openworldobjectdetection}, whose framework relies on incremental learning. By incorporating an energy-based object recognizer into the detection head, the model gains the ability to identify unknown categories. However, this method depends on replay mechanisms, requiring access to historical task data to update the model. Additionally, it often exhibits a bias toward known categories when handling unknown objects, limiting its generalization capabilities.
To address these limitations, many subsequent studies have been proposed. For instance, ~\cite{revisiting,ma2023annealing} improved the experimental setup for OWOD by introducing more comprehensive benchmark datasets and stricter evaluation metrics, enhancing the robustness of unknown category detection. While these improvements achieved promising results in controlled experimental settings, their adaptability to complex scenarios and dynamic category changes remains inadequate.
Recent research has shifted focus toward optimizing the feature space to better separate known and unknown categories. 
Methods such as ~\cite{Sun_2024_CVPR, uc-owod, fang2023unsupervisedrecognitionunknownobjects, Yu_2022} propose advancements in feature space extraction, enabling models to more effectively extract feature information for the localization and identification of unknown objects. Recently, several methods ~\cite{fomo, skdf, fromovdtoowd} have emerged, leveraging pretrained models for open-world object detection and achieving significant improvements.

\subsection{Parameter Efficient Learning}
Prompt learning has emerged as a significant research direction in both natural language processing (NLP) and computer vision. By providing carefully designed prompts to pre-trained large models such as ~\cite{clip}, prompt learning enables models to perform specific tasks in unsupervised or semi-supervised settings efficiently. Methods such as ~\cite{coop,cocoop,maple,xiong2025pyra,yang2024promptable,hao2023consolidatormergeableadaptergrouped} introduce learnable prompt embeddings, moving beyond fixed, handcrafted prompts to enhance flexibility across various visual downstream tasks. And DetPro ~\cite{detpro} is the first to apply it to open-vocabulary object detection, achieving significant improvements using learnable prompts derived from text inputs.

Low-Rank Adaptation (LoRA)~\cite{lora2021} and its derivatives~\cite{zeng2023expressive,dora,cliplora}, as a parameter-efficient fine-tuning technique, has demonstrated outstanding performance in adapting large models. By inserting trainable low-rank decomposition modules into the weight matrices of pre-trained models without altering the original weights, LoRA significantly reduces the number of trainable parameters. CLIP-LoRA~\cite{cliplora} introduces LoRA into VLM models as a replacement for adapters and prompts, enabling fine-tuning for downstream tasks with faster training speeds and improved performance. 

%% file: sec/3_Method.tex
\section{Efficient Universal Open-World Object Detection}
\label{sec:method}

\subsection{Problem Definition}
\textbf{Universal Open-World Object Detection (Uni-OWD)} extends the challenges of Open Vocabulary Detection (OVD) and Open-World Object Detection (OWOD), aiming to create a unified framework that not only detects known objects in the vocabulary but also dynamically adapts to unknown objects while maintaining scalability and efficiency in real-world scenarios.

Define the object category set as $\mathcal{C} = \mathcal{C}_{\text{k}} \cup \mathcal{C}_{\text{unk}}$, where $\mathcal{C}_{\text{k}}$ represents the set of known categories, $\mathcal{C}_{\text{unk}}$ represents the set of unknown categories, and $\mathcal{C}_{\text{k}} \cap \mathcal{C}_{\text{unk}} = \varnothing$. Given an input image $\mathcal{I}$ and a vocabulary $\mathcal{V}$, the goal of Uni-OWD is to design a detector $\mathcal{D}$ that satisfies the following objectives:

\begin{enumerate}
    \item For each category $c_k \in \mathcal{C}_{\text{k}}$, represented by its text $\mathcal{T}_{c_k}$ $\in \mathcal{V}$, the detector $\mathcal{D}$ should accurately predict the bounding boxes $\mathcal{B}_{c_k}$ and their associated category labels $c_k$ by $\mathcal{D}(\mathcal{I}, \mathcal{V}) \rightarrow \{(b, c_k) \mid b \in \mathcal{B}_{c_k}, c_k \in \mathcal{C}_{\text{k}}\}$

    \item For objects belonging to $\mathcal{C}_{\text{unk}}$, the detector should identify their bounding boxes $\mathcal{B}_{\text{unk}}$ and assign them the generic label ``unknown" with a wildcard $\mathcal{T}_w$, such that: $
        \mathcal{D}(\mathcal{I}, \mathcal{T}_w) \rightarrow \{(b, \text{unknown}) \mid b \in \mathcal{B}_{\text{unk}}\}$

    \item The detector can iteratively expand the known category set $\mathcal{C}_{\text{k}}$ and vocabulary $\mathcal{V}$ by discovering new categories $\mathcal{C}_{\text{new}}$ from $\mathcal{C}_{\text{unk}}$, represented as $\mathcal{C}_{\text{k}}^{t+1} = \mathcal{C}_{\text{k}}^t \cup \mathcal{C}_{\text{new}}$
\end{enumerate}

The Uni-OWD framework is designed to develop a detector that leverages a textual vocabulary and a wildcard to identify both known and unknown object categories within an image, combining the strengths of open-vocabulary and open-world detection tasks. It ensures precise detection and classification for known categories while assigning a generic ``unknown" label to unidentified objects. This design promotes adaptability and scalability, making it well-suited for dynamic and real-world applications.

\subsection{Efficient Adaptive Decision Learning}

Designing a universal open-world object detection model suitable for deployment on edge and mobile devices demands a strong emphasis on efficiency. Traditional open-vocabulary detection models \cite{glipv2, groundingdino, yoloworld, grounding15} align text and image modalities by introducing fine-grained fusion operations in the early layers.  Then they rely on contrastive learning for both modalities to establish decision boundaries for object classification, enabling the model to adapt dynamically to novel classes during inference by leveraging new textual inputs.

YOLO-World \cite{yoloworld} proposed an efficient architecture, RepVL-PAN, to perform image-text fusion through re-parameterization. Despite its advancements, the model's inference speed is still heavily influenced by the number of textual class inputs. This poses a challenge for low-compute devices, where performance degrades sharply as the number of text inputs increases, making it unsuitable for real-time detection tasks in complex, multi-class scenarios. To address this, we propose a adaptive decision learning strategy (AdaDL) to eliminate the heavy early-layer fusion operation.

During the construction of decision boundaries, most existing methods freeze the text encoder and rely on pre-trained models, such as BERT\cite{BERT2019} or CLIP\cite{clip}, to extract textual features for interaction with visual features. Without a fusion structure, the text features struggle to capture image-related information dynamically, leading to suboptimal multimodal decision boundary construction when adjustments are made solely to the image features. To overcome this, our AdaDL strategy aims to enhance the decision representation during training for the Uni-OWD scenario. Specifically, during training, we introduce efficient parameters into the text encoder by incorporating Low-Rank Adaptation (LoRA) into all query, key, value and output projection layers, which can be described as:
\begin{equation}
    \label{eq:text_encoder}
    h=W'x=W_0x + \Delta Wx
\end{equation}
where $W_0$ represents the pretrained weights of the CLIP text encoder, and $\Delta W$ is the product of two low-rank matrices. The model's input and output are $x$ and $h$. The rank is set to a value much smaller than the model’s feature dimension. 
This strategy ensures that the pre-trained parameters of the text encoder remain unchanged while low-rank matrices dynamically store information related to cross-modality interactions during training. By continuously calibrating the outputs of text encoder, this method allows the decision boundaries constructed by both modalities to adapt more effectively to each other. In practice, the calibrated text embeddings can be precomputed and stored offline during inference, thereby avoiding the computational cost of the text encoder.

\textbf{YOLOv10 as the efficient object detector.} To improve the efficiency, we accommodate the proposed adaptive decision learning strategy into the recent advanced YOLOv10 \cite{yolov10} as the efficient object detector.  We employ a multimodal dual-head match to adapt the decision boundary for both classification heads in YOLOv10. Specifically, during the region-text contrastive learning between the region anchor and the class text, we refine the region embeddings from two heads by aligning them with shared, semantically rich text representations, enabling seamless end-to-end training and inference. Furthermore, we integrate a consistent dual alignment strategy for region contrastive learning, where the dual-head matching process is formalized as:
\begin{equation}
    \label{eq:label-assignment}
    m(\alpha, \beta) = s^{\alpha } \times u^{\beta }
\end{equation}
where $u$ represents the IoU value between the predicted box and the ground truth box. $s$ is the classification score obtained by multi-modal information, which is derived as:
\begin{equation}
    s = \text{sim}(I,T)
\end{equation}
where $\text{sim}(\cdot,\cdot)$ is the cosine similarity, $T$ is the embeddings from the text $\mathcal{T} \in \mathcal{V}$ and $I$ is the pixel-level features from image $\mathcal{I}$. To ensure minimal supervision gap between the both heads during multimodel dual-head matching, we adopt the consistent settings, where $\alpha_{o2o} = \alpha_{o2m}$ and $\beta_{o2o} = \beta_{o2m}$. This allows the one-to-one head to effectively learn consistent supervisory signals with one-to-many head.

As a result, the calibrated text encoder and YOLO structure can operate entirely independently in the early stages, eliminating the need for fusion operations while efficiently adapting to better multimodal decision boundaries.

\begin{figure}
    \centering
    \includegraphics[width=\linewidth]{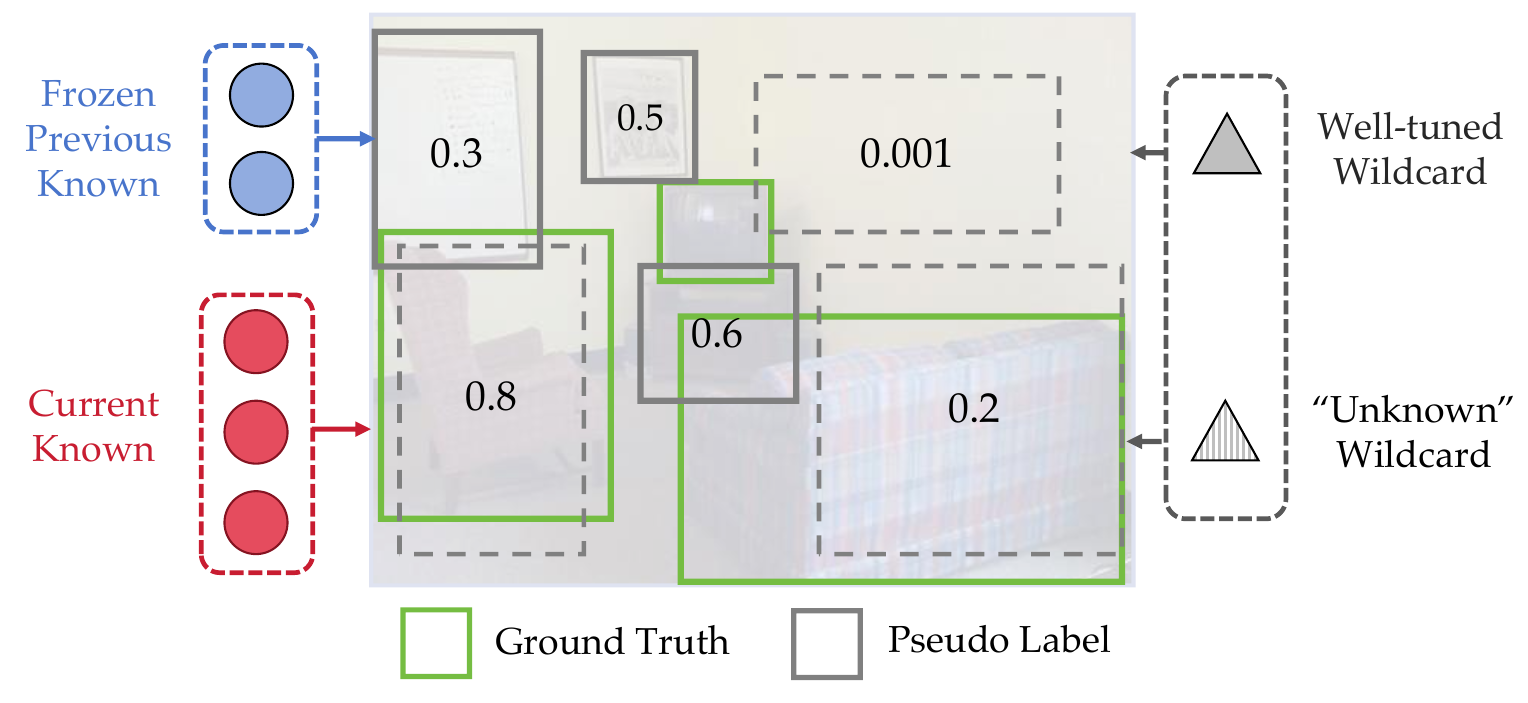}
    \caption{\textbf{The Process of Known/Wildcard Learning.} The text embeddings for previously known classes are frozen, while the embeddings for currently known classes are fine-tuned using ground truth labels. The ``unknown" wildcard is supervised by pseudo labels generated by the well-tuned wildcard predictions. It shows well-tuned wildcard's prediction scores and the boxes with low confidence scores or high IoU values (dashed boxes) with known class ground truth are filtered out.}
    \label{fig:wildcard}
\end{figure}

\subsection{Open-World Wildcard Learning}

In the previous section, we introduced the \textit{AdaDL} to improve the efficiency of open-vocabulary object detection, mitigating the impact of large input class text on inference latency meanwhile improves its performance. This strategy enables real-world applications to expand the vocabulary while maintaining high efficiency, covering as many objects as possible. However, open-vocabulary models inherently rely on predefined vocabularies to detect and classify objects, which limits their capability in real-world scenarios. Some objects are difficult to predict or describe using textual inputs, making it challenging for open-vocabulary models to detect these out-of-vocabulary instances.

To address this, we propose a \textit{wildcard learning} approach that enables the model to detect objects not present in the vocabulary and label them as ``unknown" rather than ignoring them. Specifically, we directly leverage a wildcard embedding to unlock generic power of open-vocabualry model. As shown in \cref{tab:ablation-owod}, after the decision adaptation, the wildcard $\mathcal{T}_w$ (e.g. ``object'') demonstrates remarkable capability in capturing unknown objects within a scene in a zero-shot manner. To further enhance its effectiveness, we fine-tune its text embedding on the pretraining dataset for a few epochs. During this process, all ground-truth instances are treated as belonging to the same ``object" class. This fine-tuning enables the embedding to capture richer semantics, empowering the model to identify objects that might have been overlooked by the predefined specific classes. 

To avoid duplicate predictions for known classes, we utilize this well-tuned wildcard embedding $T_{obj}$ to teach an ``unknown'' wildcard embedding $T_{unk}$. The ``unknown'' wildcard is trained in a self-supervised manner without relying on ground-truth labels of ``unknown'' class. As shown in \cref{fig:wildcard}, predictions that has the highest similarity score with $T_{obj}$ across all known classes embeddings are used as pseudo label candidates. To further refine these candidates, we introduce a simple selection process:
\begin{equation}
    \label{eq:pseudo-select}
        \Phi(s, u) = \left\{\begin{matrix} 1,& \quad  \text{if} \space (u < \sigma_1) \space 
    \wedge  (\space s > \sigma_2 )
     \\
    0,& \quad \text{otherwise}
    \end{matrix}\right.
\end{equation}
where $u$ is the maximum IoU between predictions and known class ground truth boxes. And the predictions with $u$ below a threshold $\sigma_1$ or classification score $s$ above a threshold $\sigma_2$ are selected. The remaining predictions are assigned to $T_{unk}$ as target labels.

For known classes, only their corresponding text embeddings $T_k$ from $\mathcal{T}_{k} \in \mathcal{V}$ are fine-tuned on downstream tasks by multimodal dual-head match to enhance their similarity scores $s_k$ aligned with target score $o_k$. These embeddings are subsequently frozen to preserve performance and avoid degradation when new classes are introduced. Unlike traditional open-world methods ~\cite{joseph2021openworldobjectdetection, owdetr} relying on exemplar replay to do incremental learning, our method can avoid catastrophic forgetting without extra exemplars, as each text embeddings are fine-tuned independently.

Since $T_k$, $T_{obj}$ and $T_{unk}$ calculate similarity scores only in frozen classification head, there is no loss from box regression, focusing exclusively on learning class-specific information. The soft target scores of $T_{unk}$ are directly derived from the similarity scores $s_{obj}$ from the $T_{obj}$. Therefore, the fine-tuning loss is formulated as the combination of the current known loss and the unknown loss, ensuring the model learns effectively from both known and unknown categories during training:
\begin{equation}
    \label{eq:fine-tuning}
    \mathcal{L} = \mathcal{L}_{k} (s_k, o_k) + \Phi(s_{obj}, u_{obj}) \cdot \mathcal{L}_{unk} (s_{unk}, s_{obj})
\end{equation}
where $s_{unk}$ is the prediction score from ``unknown'' wildcard. $\mathcal{L}$ represents the binary cross-entropy (BCE) loss. During inference, we employ a simple and efficient unknown filtering strategy $\mathcal{F}$ for unknown class predictions $P_{unk}$ that have a high IoU with confident known class predictions $P_k$ to further de-duplicate.
\begin{small}
\begin{equation}
    \label{eq:unk-filter}
    \mathcal{F}(P_{unk})=\left \{ p_u \in P_{unk} \mid \text{IoU}(p_u, p_k) < \tau, \forall p_k \in P_k \right \}
\end{equation}
\end{small}
where $\tau$ is the IoU threshold for unknown filtering. 

Subsequently, new categories can be discovered from the predictions of the unknown class, and their class names will be added to the vocabulary $\mathcal{V}$, where they serve as known classes for the next iteration.

\begin{table*}[t]
\centering
\caption{\textbf{Zero-shot evaluation on the LVIS minival dataset.} The AP of our model is presented for both the one-to-one head (left) and the one-to-many head (right). All speed measurements are conducted on a V100 GPU using PyTorch without TensorRT. FPS$^f$ denotes the speed in the forward process without post-processing. We limit number of prediction boxes under 3000 before NMS for FPS measurement.}
\label{tab:lvis-eval}
\large 
\setlength{\tabcolsep}{6pt} 
\renewcommand{\arraystretch}{1.3} 
\resizebox{\linewidth}{!}{%
\begin{tabular}{cccccccccc} 
\toprule
Model                   & Backbone        & Pre-trained Data             & Params & $AP$        & $AP_r$       & $AP_c$       & $AP_f$       & $FPS$  & $FPS^f$  \\ 
\midrule
GLIPv2-T                & Swin-T          & O365,GoldG,Cap4M & 232M   & 29.0      & -         & -         & -         & 0.12 & -         \\
Grounding DINO 1.5 Edge & EfficientViT-L1 & Grounding-20M    & -      & 33.5      & 28.0      & 34.3      & 33.9      & -    & -         \\
OmDet-Turbo-T           & Swin-T          & O365,GoldG       & -      & 30.3      & -         & -         & -         & -    & -         \\ 
\hline
YOLO-World-S            & YOLOv8-S        & O365,GoldG       & 13M    & 24.3      & 16.6      & 22.1      & 27.7      & -    & 74.1      \\
YOLO-World-M            & YOLOv8-M        & O365,GoldG       & 29M    & 28.6      & 19.7      & 26.6      & 31.9      & -    & 58.1      \\
YOLO-World-L            & YOLOv8-L        & O365,GoldG       & 48M    & 32.5      & 22.3      & 30.6      & 36.1      & -    & 52.0      \\
YOLO-Worldv2-S          & YOLOv8-S        & O365,GoldG       & 13M    & 22.7      & 16.3      & 20.8      & 25.5      & 87.3 & 113.4     \\
YOLO-Worldv2-M          & YOLOv8-M        & O365,GoldG       & 29M    & 30.0      & 25.0      & 27.2      & 33.4      & 74.6 & 90.3      \\
YOLO-Worldv2-L          & YOLOv8-L        & O365,GoldG       & 48M    & 33.0      & 22.6      & 32.0      & 35.8      & 51.2 & 58.6      \\ 
\hline
\textbf{YOLO-UniOW-S}         & \textbf{YOLOv10-S}       & \textbf{O365,GoldG}       & \textbf{7.5M}   & \textbf{26.2/27.4} & \textbf{24.1/26.0} & \textbf{24.9/25.6} & \textbf{27.7/29.3} & \textbf{98.3} & \textbf{119.3}     \\
\textbf{YOLO-UniOW-M}         & \textbf{YOLOv10-M}       & \textbf{O365,GoldG}       & \textbf{16.2M}  & \textbf{31.8/32.8} & \textbf{26/26.6}   & \textbf{30.5/31.8} & \textbf{34/34.9}   & \textbf{86.2} & \textbf{98.9}      \\
\textbf{YOLO-UniOW-L}         & \textbf{YOLOv10-L}       & \textbf{O365,GoldG}       & \textbf{29.4M}  &       \textbf{34.6/34.8}	  &     \textbf{30/34.2}      &     \textbf{33.6/32.4}      &     \textbf{36.3/37.0}      & \textbf{64.8} & \textbf{69.6}      \\
\bottomrule
\end{tabular}%
}
\end{table*}

%% file: sec/4_Experiments.tex
\section{Experiments}
\label{sec:experiments}

\subsection{Dataset}
We evaluate our method on two distinct setups, targeting both OVD and OWOD. Our experiments leverage diverse datasets to comprehensively assess the model's performance in detecting known and unknown objects.

\textbf{Open-Vocabulary Object Detection:}
For open vocabulary detection, the model is trained on a combination of Objects365~\cite{object365} and GoldG~\cite{GoldG} datasets, and evaluated on the LVIS~\cite{livs} dataset.The LVIS dataset contains 1,203 categories, exhibiting a realistic long-tailed distribution of rare, common, and frequent classes. This setup focuses on evaluating the model's capacity to align visual and language representations, detect novel and unseen categories, and generalize across a large-scale, long-tailed dataset.

\textbf{Open-World Object Detection:}
For open-world object detection, we evaluate our method on three established OWOD benchmarks: \textbf{M-OWODB}: This benchmark combines the COCO~\cite{coco} and PASCAL VOC~\cite{VOC} datasets, where known and unknown classes are mixed across tasks. It is divided into four sequential tasks. At each task, the model learns new classes while the remaining classes remain unknown. \textbf{S-OWODB}: Based solely on COCO, this benchmark separates known and unknown classes by their superclass. \textbf{nu-OWODB}: This benchmark is derived from~\cite{fromovdtoowd} and based on the nuSences dataset~\cite{nuScenes}.This benchmark is specifically designed to evaluate the model's capability in autonomous driving scenarios. The nu-OWODB captures the complexity of urban driving environments, including crowded city streets, challenging weather conditions, frequent occlusions, and dense traffic with intricate interactions between objects. 

By incorporating these benchmarks, we assess the model's ability to handle real-world OWOD challenges while maintaining robustness and scalability across diverse settings.

\subsection{Evaluation Metrics}

\textbf{Open-Vocabulary Evaluation:}
Similar to YOLO-World and other pre-trained models, we evaluate the zero-shot capability of our pre-trained model on the LVIS minival dataset, which contains the same images in COCO validation set. For fair and consistent comparison, we use \textit{standard AP} metrics to measure the model's performance.

\textbf{Open-World Evaluation:}
We adapt the pretrained open-vocabulary model to the open-world scenario, enabling it to recognize both known and unknown objects. For known objects, we use \textit{mAP} as the evaluation metric. To further assess catastrophic forgetting during incremental tasks, the mAP is divided into previous known (PK) and current known (CK) categories.
For unknown objects, since it is impractical to exhaustively annotate all remaining objects in the scene, we employ the \textit{Recall} metric to evaluate the model's ability to detect unknown categories. Additionally, WI~~\cite{joseph2021openworldobjectdetection} and A-OSE~~\cite{joseph2021openworldobjectdetection} are used to measure the extent to which unknown objects interfere with known object predictions. However, due to their instability, these metrics are provided for reference purposes only.

\begin{table*}[t]
\centering
%\label{tab:owod-eval}
\caption{\textbf{OWOD results on M-OWODB (top) and S-OWODB (bottom).} Comparision of unknown class recall (U-Recall) and mean average precision (mAP) for known classes. Our method outperforms both traditional models and those leveraging pretrained knowledge. OVOW* represents the reproduced version using YOLO-Worldv2-S to ensure a fair comparison with our model at the same scale.}
\resizebox{16.5cm}{!}{
\setlength{\extrarowheight}{0pt}
\addtolength{\extrarowheight}{\aboverulesep}
\addtolength{\extrarowheight}{\belowrulesep}
\setlength{\aboverulesep}{0pt}
\setlength{\belowrulesep}{0pt}
\begin{tabular}{c|cc|cccc|cccc|ccc} 
\toprule
Task IDs (→)             & \multicolumn{2}{c|}{Task1}                                                                             & \multicolumn{4}{c|}{Task 2}                                                                                                 & \multicolumn{4}{c|}{Task3}                                                                                                  & \multicolumn{3}{c}{Task 4}                                     \\ 
\hline
\multirow{2}{*}{Methods} & {\cellcolor[rgb]{1,1,0.918}}                               & {\cellcolor[rgb]{0.918,0.961,1}}mAP (↑) & {\cellcolor[rgb]{1,1,0.918}}                               & \multicolumn{3}{c|}{{\cellcolor[rgb]{0.918,0.961,1}}mAP (↑)} & {\cellcolor[rgb]{1,1,0.918}}                               & \multicolumn{3}{c|}{{\cellcolor[rgb]{0.918,0.961,1}}mAP (↑)} & \multicolumn{3}{c}{{\cellcolor[rgb]{0.918,0.961,1}}mAP (↑)}  \\
                         & \multirow{-2}{*}{{\cellcolor[rgb]{1,1,0.918}}U-Recall (↑)} & CK                                        & \multirow{-2}{*}{{\cellcolor[rgb]{1,1,0.918}}U-Recall (↑)} & PK    & CK    & Both                                           & \multirow{-2}{*}{{\cellcolor[rgb]{1,1,0.918}}U-Recall (↑)} & PK    & CK    & Both                                           & PK    & CK    & Both                                           \\ 
\hline
\multicolumn{14}{c}{M-OWODB}                                                                                                                                                                                                                                                                                                                                                                                                                                    \\ 
\hline
ORE-EBUI~\cite{joseph2021openworldobjectdetection}                 & {\cellcolor[rgb]{1,1,0.918}}4.9~                           & 56.0~                                     & {\cellcolor[rgb]{1,1,0.918}}2.9~                           & 52.7~ & 26.0~ & 39.4~                                          & {\cellcolor[rgb]{1,1,0.918}}3.9~                           & 38.2~ & 12.7~ & 29.7~                                          & 29.6~ & 12.4~ & 25.3~                                          \\
OW-DETR~\cite{owdetr}                  & {\cellcolor[rgb]{1,1,0.918}}7.5~                           & 59.2~                                     & {\cellcolor[rgb]{1,1,0.918}}6.2~                           & 53.6~ & 33.5~ & 42.9~                                          & {\cellcolor[rgb]{1,1,0.918}}5.7~                           & 38.3~ & 15.8~ & 30.8~                                          & 31.4~ & 17.1~ & 27.8~                                          \\
PROB~\cite{prob}                     & {\cellcolor[rgb]{1,1,0.918}}19.4~                          & 59.5~                                     & {\cellcolor[rgb]{1,1,0.918}}17.4~                          & 55.7~ & 32.2~ & 44.0~                                          & {\cellcolor[rgb]{1,1,0.918}}19.6~                          & 43.0~ & 22.2~ & 36.0~                                          & 35.7~ & 18.9~ & 31.5~                                          \\
CAT~\cite{cat}                      & {\cellcolor[rgb]{1,1,0.918}}23.7~                          & 60.0~                                     & {\cellcolor[rgb]{1,1,0.918}}19.1~                          & 55.5~ & 32.2~ & 44.1~                                          & {\cellcolor[rgb]{1,1,0.918}}24.4~                          & 42.8~ & 18.8~ & 34.8~                                          & 34.4~ & 16.6~ & 29.9~                                          \\
RandBox~\cite{randbox}                  & {\cellcolor[rgb]{1,1,0.918}}10.6                           & 61.8                                      & {\cellcolor[rgb]{1,1,0.918}}6.3                            & -     & -     & 45.3                                           & {\cellcolor[rgb]{1,1,0.918}}7.8                            & -     & -     & 39.4                                           & -     & -     & 35.4                                           \\
EO-OWOD~\cite{eo-owod}                  & {\cellcolor[rgb]{1,1,0.918}}24.6                           & 61.3                                      & {\cellcolor[rgb]{1,1,0.918}}26.3                           & 55.5  & 38.5  & 47                                             & {\cellcolor[rgb]{1,1,0.918}}29.1                           & 46.7  & 30.6  & 41.3                                           & 42.4  & 24.3  & 37.9                                           \\
MEPU-FS~\cite{mepu-fs}                  & {\cellcolor[rgb]{1,1,0.918}}31.6~                          & 60.2~                                     & {\cellcolor[rgb]{1,1,0.918}}30.9~                          & 57.3~ & 33.3~ & 44.8~                                          & {\cellcolor[rgb]{1,1,0.918}}30.1~                          & 42.6~ & 21.0~ & 35.4~                                          & 34.8~ & 18.9~ & 30.4~                                          \\
MAVL~\cite{mavl}                     & {\cellcolor[rgb]{1,1,0.918}}50.1~                          & 64.0~                                     & {\cellcolor[rgb]{1,1,0.918}}49.5~                          & 61.6~ & 30.8~ & 46.2~                                          & {\cellcolor[rgb]{1,1,0.918}}50.9~                          & 43.8~ & 22.7~ & 36.8~                                          & 36.2~ & 20.6~ & 32.3~                                          \\
SKDF~\cite{skdf}                     & {\cellcolor[rgb]{1,1,0.918}}39.0~                          & 56.8~                                     & {\cellcolor[rgb]{1,1,0.918}}36.7~                          & 52.3~ & 28.3~ & 40.3~                                          & {\cellcolor[rgb]{1,1,0.918}}36.1~                          & 36.9~ & 16.4~ & 30.1~                                          & 31.0~ & 14.7~ & 26.9~                                          \\
OVOW*~\cite{fromovdtoowd}                    & {\cellcolor[rgb]{1,1,0.918}}65.9                          & 57.1                                     & {\cellcolor[rgb]{1,1,0.918}}72.7                         & 55.1 & 38.6 & 47.0                                          & {\cellcolor[rgb]{1,1,0.918}}66.48                          & 43.4 & 24.3 & 37.0                                          & 36.2 & 20.4 & 32.30                                          \\ \hline
\textbf{YOLO-UniOW-S}                   & {\cellcolor[rgb]{1,1,0.918}}\textbf{80.6}                          & \textbf{70.4}                                     & {\cellcolor[rgb]{1,1,0.918}}\textbf{80.8}                          & \textbf{70.4} & \textbf{42.9} & \textbf{56.6}                                          & {\cellcolor[rgb]{1,1,0.918}}\textbf{79.9}                          & \textbf{56.6} & \textbf{33.1} & \textbf{48.8}                                          & \textbf{48.8} & \textbf{26.7} & \textbf{43.3}                                          \\
\textbf{YOLO-UniOW-M}                   & {\cellcolor[rgb]{1,1,0.918}}\textbf{82.6}                          & \textbf{73.6}                                     & {\cellcolor[rgb]{1,1,0.918}}\textbf{82.6}                          & \textbf{73.4} & \textbf{48.4} & \textbf{60.9}                                          & {\cellcolor[rgb]{1,1,0.918}}\textbf{81.5}                          & \textbf{60.9} & \textbf{39.0} & \textbf{53.6}                                          & \textbf{53.6} & \textbf{32.0} & \textbf{48.2}                                          \\ 
\hline\hline
\multicolumn{14}{c}{S-OWODB}                                                                                                                                                                                                                                                                                                                                                                                                                                    \\ 
\hline
ORE-EBUI~\cite{joseph2021openworldobjectdetection}                 & {\cellcolor[rgb]{1,1,0.918}}1.5~                           & 61.4~                                     & {\cellcolor[rgb]{1,1,0.918}}3.9~                           & 56.7~ & 26.1~ & 40.6~                                          & {\cellcolor[rgb]{1,1,0.918}}3.6~                           & 38.7~ & 23.7~ & 33.7~                                          & 33.6~ & 26.3~ & 31.8~                                          \\
OW-DETR~\cite{owdetr}                  & {\cellcolor[rgb]{1,1,0.918}}5.7~                           & 71.5~                                     & {\cellcolor[rgb]{1,1,0.918}}6.2~                           & 62.8~ & 27.5~ & 43.8~                                          & {\cellcolor[rgb]{1,1,0.918}}6.9~                           & 45.2~ & 24.9~ & 38.5~                                          & 38.2~ & 28.1~ & 33.1~                                          \\
PROB~\cite{prob}                     & {\cellcolor[rgb]{1,1,0.918}}17.6~                          & 73.4~                                     & {\cellcolor[rgb]{1,1,0.918}}22.3~                          & 66.3~ & 36.0~ & 50.4~                                          & {\cellcolor[rgb]{1,1,0.918}}24.8~                          & 47.8~ & 30.4~ & 42.0~                                          & 42.6~ & 31.7~ & 39.9~                                          \\
CAT~\cite{cat}                      & {\cellcolor[rgb]{1,1,0.918}}24.0~                          & 74.2~                                     & {\cellcolor[rgb]{1,1,0.918}}23.0~                          & 67.6~ & 35.5~ & 50.7~                                          & {\cellcolor[rgb]{1,1,0.918}}24.6~                          & 51.2~ & 32.6~ & 45.0~                                          & 45.4~ & 35.1~ & 42.8~                                          \\
EO-OWOD~\cite{eo-owod}                  & {\cellcolor[rgb]{1,1,0.918}}24.6                           & 71.6                                      & {\cellcolor[rgb]{1,1,0.918}}27.9                           & 64    & 39.9  & 51.3                                           & {\cellcolor[rgb]{1,1,0.918}}31.9                           & 52.1  & 42.2  & 48.8                                           & 48.7  & 38.8  & 46.2                                           \\
MEPU-FS~\cite{mepu-fs}                  & {\cellcolor[rgb]{1,1,0.918}}37.9~                          & 74.3~                                     & {\cellcolor[rgb]{1,1,0.918}}35.8~                          & 68.0~ & 41.9~ & 54.3~                                          & {\cellcolor[rgb]{1,1,0.918}}35.7~                          & 50.2~ & 38.3~ & 46.2~                                          & 43.7~ & 33.7~ & 41.2~                                          \\
SKDF~\cite{skdf}                     & {\cellcolor[rgb]{1,1,0.918}}60.9~                          & 69.4~                                     & {\cellcolor[rgb]{1,1,0.918}}60.0~                          & 63.8~ & 26.9~ & 44.4~                                          & {\cellcolor[rgb]{1,1,0.918}}58.6~                          & 46.2~ & 28.0~ & 40.1~                                          & 41.8~ & 29.6~ & 38.7~                                          \\
OVOW*~\cite{fromovdtoowd}                    & {\cellcolor[rgb]{1,1,0.918}}69.8~                          & 52.7~                                     & {\cellcolor[rgb]{1,1,0.918}}73.2~                          & 54.1~ & 37.1~ & 45.1~                                          & {\cellcolor[rgb]{1,1,0.918}}69.7~                          & 42.3~ & 31.6~ & 38.7~                                          & 38.6~ & 29.5~ & 36.4~                                          \\ \hline
\textbf{YOLO-UniOW-S}                   & {\cellcolor[rgb]{1,1,0.918}}\textbf{82.2}                           & \textbf{69.20}                                     & {\cellcolor[rgb]{1,1,0.918}}\textbf{81.4}                          & \textbf{69.2} & \textbf{50.64} & \textbf{59.4}                                          & {\cellcolor[rgb]{1,1,0.918}}\textbf{81.0}                             & \textbf{59.4} & \textbf{44.4} & \textbf{54.4}                                          & \textbf{54.4} & \textbf{44.7} & \textbf{52.0}                                          \\
\textbf{YOLO-UniOW-M}                   & {\cellcolor[rgb]{1,1,0.918}}\textbf{84.5}                          & \textbf{74.4}                                     & {\cellcolor[rgb]{1,1,0.918}}\textbf{83.4}                          & \textbf{74.4} & \textbf{56.9} & \textbf{65.2}                                          & {\cellcolor[rgb]{1,1,0.918}}\textbf{83.0}                           & \textbf{65.2} & \textbf{52.2} & \textbf{61.0}                                          & \textbf{61.0} & \textbf{52.7} & \textbf{58.9}                                          \\
\bottomrule
\end{tabular}
}
\label{tab:owod-eval}
\end{table*}

\subsection{Implementation Details}

\textbf{Open-Vocabulary Detection:}
Our image detector follows YOLOv10~\cite{yolov10}, which provides an efficient design for dual-head training. Similar to YOLO-World~\cite{yoloworld}, we utilize a pre-trained CLIP text encoder. However, we do not perform image-text fusion in the neck. Instead, we align the two modalities solely in the head using efficient adaptive decision learning.
During pretraining, we incorporate low-rank matrices into the all projection layers in the CLIP text encoder. The rank for the matrices are set to 16. Our pretraining is conducted on 8 GPUs, with a batch size of 128. Both the YOLO model and LoRA parameters for the text encoder are trained with an initial learning rate of $5 \times 10^{-4} $ and weight decay of 0.025.

\textbf{Open-World Detection:}
All the wildcard embeddings are initialized from text features extracted from a generic text, ``object", by our calibrated text encoder. We use the same training datasets employed for open-vocabulary pretraining to fine-tune the wildcard embedding $T_{obj}$. Specifically, the wildcard embedding is trained for 3 epochs with a learning rate of $ 1 \times 10^{-4} $. Using the well-tuned wildcard as an anchor, the learning rate for fine-tuning the known and unknown class embeddings is set to $ 1 \times 10^{-3} $, with weight decay set to 0. All the other parts of model are frozen and mosaic augmentation is not applied during this stage.  

For training the ``unknown'' wildcard, pseudo-labels are selected based on an IoU threshold $ \sigma_1 = 0.5 $ and a score threshold $ \sigma_2 = 0.01 $. During inference, known class predictions with score greater than 0.2 are confident predictions, and $\tau=0.99$. For known class detection, predictions with scores below 0.05 are filtered out as default.

All fine-tuning experiments are conducted on 8 GPUs, with a batch size of 16 per GPU. Notably, all open-world experiments are evaluated using the one-to-one head, which not requires NMS operations for post-processing.

\begin{table*}[t]
\caption{\textbf{Evaluation on nu-OWODB.} Our method outperforms all other approaches in unknown metrics, demonstrating its strong adaptability for real-world applications. OVOW* represents the reproduced model based on YOLO-Worldv2-S }
\label{tab:nuowod-eval}
\centering
\resizebox{16.5cm}{!}{
\setlength{\extrarowheight}{0pt}
\addtolength{\extrarowheight}{\aboverulesep}
\addtolength{\extrarowheight}{\belowrulesep}
\setlength{\aboverulesep}{0pt}
\setlength{\belowrulesep}{0pt}
\begin{tabular}{c|cccc|cccccc|ccc} 
\toprule
Task IDs (→)             & \multicolumn{4}{c|}{Task 1}                                                                                                                                                                                                                                   & \multicolumn{6}{c|}{Task 2}                                                                                                                                                                                                                                                        & \multicolumn{3}{c}{Task3}                                    \\ 
\hline
\multirow{2}{*}{Methods} & {\cellcolor[rgb]{1,1,0.918}}                         & \multicolumn{1}{l}{{\cellcolor[rgb]{1,1,0.918}}}                            & \multicolumn{1}{l}{{\cellcolor[rgb]{1,1,0.918}}}                               & {\cellcolor[rgb]{0.918,0.961,1}}mAP (↑) & {\cellcolor[rgb]{1,1,0.918}}                         & \multicolumn{1}{l}{{\cellcolor[rgb]{1,1,0.918}}}                            & \multicolumn{1}{l}{{\cellcolor[rgb]{1,1,0.918}}}                               & \multicolumn{3}{c|}{{\cellcolor[rgb]{0.918,0.961,1}}mAP (↑)} & \multicolumn{3}{c}{{\cellcolor[rgb]{0.918,0.961,1}}mAP (↑)}  \\
                         & \multirow{-2}{*}{{\cellcolor[rgb]{1,1,0.918}}WI (↓)} & \multicolumn{1}{l}{\multirow{-2}{*}{{\cellcolor[rgb]{1,1,0.918}}A-OSE (↓)}} & \multicolumn{1}{l}{\multirow{-2}{*}{{\cellcolor[rgb]{1,1,0.918}}U-Recall (↑)}} & CK                                      & \multirow{-2}{*}{{\cellcolor[rgb]{1,1,0.918}}WI (↓)} & \multicolumn{1}{l}{\multirow{-2}{*}{{\cellcolor[rgb]{1,1,0.918}}A-OSE (↓)}} & \multicolumn{1}{l}{\multirow{-2}{*}{{\cellcolor[rgb]{1,1,0.918}}U-Recall (↑)}} & PK    & CK    & Both                                         & PK    & CK    & Both                                         \\ 
\hline
PROB~\cite{prob}                     & 0.0025                             & 2897                                  & 0.5                                      & 25.1                                      & 0.0015                             & 1583                                  & 2.8                                      & 27.2  & 6.7   & 18.8                                           & 18.1  & 16    & 17.5                                           \\
EO-OWOD~\cite{eo-owod}                  & 0.0059                             & 223                                   & 1.4                                      & 22.4                                      & 0.003                              & 172                                   & 0.8                                      & 27    & 13.5  & 21.4                                           & 21.8  & 25.6  & 22.8                                           \\
OVOW*~\cite{fromovdtoowd}                    & 0.0080                             & 16478                                 & 16.7                                   & 14.2                                    & 0.0096                            & 6394                                  & 21.74                                   & 13.6 & 6.0  & 10.5                                          & 10.0  & 9.7  & 9.9                                         \\ \hline
\textbf{YOLO-UniOW-S}                   & \textbf{0.0137}                            & \textbf{1658}                                  & \textbf{37.5}                                    & \textbf{21.5}                                     & \textbf{0.0074}                            & \textbf{1265}                                  & \textbf{30.0}                                    & \textbf{21.5} & \textbf{9.8}  & \textbf{16.7}                                          & \textbf{16.7} & \textbf{15.6} & \textbf{16.4}                                          \\
\textbf{YOLO-UniOW-M}                   & \textbf{0.0147}                            & \textbf{1722}                                  & \textbf{41.8}                                    & \textbf{24.8}                                     & \textbf{0.0067}                            & \textbf{1156}                                  & \textbf{35.4}                                    & \textbf{24.8} & \textbf{13.1} & \textbf{20.0}                                             & \textbf{20.0}    & \textbf{18.4}  & \textbf{19.6}                                          \\
\bottomrule
\end{tabular}}
\end{table*}

\begin{table*}[t]
\centering
\large
\caption{\textbf{Comparison with Oracle and Zero-shot Settings.} We compare our method with the closed-set YOLOv10 model trained in an oracle manner, and our pre-trained open-vocabulary model in a zero-shot setting. Our method achieves great improvement, even surpassing the close-set oracle model.}
\label{tab:ablation-owod}
\resizebox{\linewidth}{!}{
\setlength{\extrarowheight}{0pt}
\addtolength{\extrarowheight}{\aboverulesep}
\addtolength{\extrarowheight}{\belowrulesep}
\setlength{\aboverulesep}{0pt}
\setlength{\belowrulesep}{0pt}
\begin{tabular}{c|cccc|cccccc|cccccc} 
\toprule
Task IDs (→)             & \multicolumn{4}{c|}{Task 1}                                                                                                                                                                                                                                   & \multicolumn{6}{c|}{Task 2}                                                                                                                                                                                                                                                        & \multicolumn{6}{c}{Task 3}                                                                                                                                                                                                                                                         \\ 
\hline
\multirow{2}{*}{Methods} & {\cellcolor[rgb]{1,1,0.918}}                         & \multicolumn{1}{l}{{\cellcolor[rgb]{1,1,0.918}}}                            & \multicolumn{1}{l}{{\cellcolor[rgb]{1,1,0.918}}}                               & {\cellcolor[rgb]{0.918,0.961,1}}mAP (↑) & {\cellcolor[rgb]{1,1,0.918}}                         & \multicolumn{1}{l}{{\cellcolor[rgb]{1,1,0.918}}}                            & \multicolumn{1}{l}{{\cellcolor[rgb]{1,1,0.918}}}                               & \multicolumn{3}{c|}{{\cellcolor[rgb]{0.918,0.961,1}}mAP (↑)} & {\cellcolor[rgb]{1,1,0.918}}                         & \multicolumn{1}{l}{{\cellcolor[rgb]{1,1,0.918}}}                            & \multicolumn{1}{l}{{\cellcolor[rgb]{1,1,0.918}}}                               & \multicolumn{3}{c}{{\cellcolor[rgb]{0.918,0.961,1}}mAP (↑)}  \\
                         & \multirow{-2}{*}{{\cellcolor[rgb]{1,1,0.918}}WI (↓)} & \multicolumn{1}{l}{\multirow{-2}{*}{{\cellcolor[rgb]{1,1,0.918}}A-OSE (↓)}} & \multicolumn{1}{l}{\multirow{-2}{*}{{\cellcolor[rgb]{1,1,0.918}}U-Recall (↑)}} & CK                                      & \multirow{-2}{*}{{\cellcolor[rgb]{1,1,0.918}}WI (↓)} & \multicolumn{1}{l}{\multirow{-2}{*}{{\cellcolor[rgb]{1,1,0.918}}A-OSE (↓)}} & \multicolumn{1}{l}{\multirow{-2}{*}{{\cellcolor[rgb]{1,1,0.918}}U-Recall (↑)}} & PK    & CK    & Both                                         & \multirow{-2}{*}{{\cellcolor[rgb]{1,1,0.918}}WI (↓)} & \multicolumn{1}{l}{\multirow{-2}{*}{{\cellcolor[rgb]{1,1,0.918}}A-OSE (↓)}} & \multicolumn{1}{l}{\multirow{-2}{*}{{\cellcolor[rgb]{1,1,0.918}}U-Recall (↑)}} & PK    & CK    & Both                                         \\ 
\hline
YOLOv10-S-oracle        & 0.0482                                             & 18447                                                                       & 61.5                                                                           & 64.6                                   & 0.0147                                              & 12614                                                                       & 84.2                                                                           & 64.3 & 48.3 & 56.3                                        & 0.0111                                              & 10795                                                                       & 83.9                                                                          & 55.9 & 31.1 & 47.6                                        \\
YOLO-UniOW-S-zs       & 0.0287                                              & 2827                                                                        & 48.8                                                                          & 68.1                                   & 0.0185                                              & 2173                                                                        & 50.5                                                                           & 68.1 & 40.1  & 54.1                                        & 0.0125                                             & 1775                                                                        & 52.1                                                                          & 53.6 & 30.7 & 46.0                                         \\
YOLO-UniOW-M-zs       & 0.0232                                              & 2409                                                                        & 54.4                                                                          & 71.0                                   & 0.0130                                              & 1740                                                                        & 57.6                                                                          & 70.8 & 44.2  & 57.5                                       & 0.00888                                              & 1448                                                                        & 59.3                                                                          & 57.7 & 37.0 & 50.8                                        \\ \hline
\textbf{YOLO-UniOW-S}                   & \textbf{0.0229}                                               & \textbf{1609}                                                                        & \textbf{80.6}                                                                          & \textbf{70.4}                                   & \textbf{0.0133}                                              & \textbf{1208}                                                                        & \textbf{80.8}                                                                          & \textbf{70.4} & \textbf{42.9} & \textbf{56.6}                                       & \textbf{0.0091}                                              & \textbf{1049}                                                                        & \textbf{79.9}                                                                          & \textbf{56.6} & \textbf{33.1} & \textbf{48.8}                                        \\
\textbf{YOLO-UniOW-M}                   & \textbf{0.0210 }                                             & \textbf{1514}                                                                        & \textbf{82.6}                                                                          & \textbf{73.4}                                   & \textbf{0.01093}                                              & \textbf{1027}                                                                        & \textbf{82.62}                                                                          & \textbf{73.4} & \textbf{48.4} & \textbf{60.9}                                        & \textbf{0.00723}                                              & \textbf{872}                                                                         & \textbf{81.5}                                                                          & \textbf{60.9} & \textbf{39.0} & \textbf{53.6}                                        \\
\bottomrule
\end{tabular}
}
\end{table*}

\subsection{Quantitative Results}

\cref{tab:lvis-eval} demonstrates that model with efficient adaptive decision learning achieves significant zero-shot performance improvements on the LVIS benchmark, outperforming recent real-time state-of-the-art open-vocabulary models~~\cite{omdet-turbo, grounding15, yoloworld}. For the small model (-S), we observe that using predictions from the one-to-one head alone improves the detection performance for rare classes by 6.4\%  and common classes by 3.2\%. Furthermore, employing a one-to-many head structure with NMS achieves even greater performance gains. This clearly demonstrates that in previous pretraining processes, the multimodal decision boundaries were fully constructed by incorporating AdaDL. Additionally, leveraging the efficient model architecture and the nature of end-to-end detection, our approach gains faster speed and eliminates the need for NMS during inference, making it highly efficient for real-world applications.

To address open-world demands, we adapt our well-adapted open-vocabulary model to recognize unknown classes that are not present in the predefined vocabulary through wildcard learning. As shown in \cref{tab:owod-eval}, the open-vocabulary model demonstrates outstanding performance in open-world scenarios due to its rich knowledge.
Through our wildcard learning strategy, the model achieves superior performance in both unknown and known class recognition compared to traditional open-world methods. Moreover, it outperforms recent open-world detection models that leverage pre-training models~~\cite{mepu-fs, skdf, fromovdtoowd}. Notably, our simpler and more efficient approach surpasses the state-of-the-art OVOW model~\cite{fromovdtoowd}, which is also based on YOLO-World structure. Our method achieves a significant improvement in unknown recall and known mAP, demonstrating its effectiveness and robustness in open-world detection tasks. Furthermore, we evaluated the model’s capability in real-world autonomous driving scenarios. As shown in \cref{tab:nuowod-eval}, our model, using a simpler approach, achieves superior unknown detection performance compared to the other methods. 

Benefiting from AdaDL and wildcard learning strategies, our model captures a broader range of unknown objects through wildcard embeddings while maintaining accurate recognition of known categories. Notably, as the model scales up, the capability of model to detect known and unknown objects increases progressively, which shows the effectiveness of our methods at different model scale.

\subsection{Ablation Study}

\textbf{Open-Vocabulary Detection:}
We conducted a series of ablation studies on the small scale model to evaluate the impact of image-text fusion. Due to differences in experimental settings, we first reproduced YOLO-Worldv2-S under our setup. Interestingly, as shown in \cref{tab:ablation-pt} our findings reveal that a smaller batch size and learning rate yield better pretraining performance, particularly improving detection for frequent classes by 2.2\%.
Building on this, we removed the VL-PAN structure and observed that the model's detection accuracy remains largely unaffected. Notably, it demonstrated improved generalization for rare classes.
Replacing YOLO-World's YOLOv8 structure with YOLOv10 and using a dual-head match demonstrated that the one-to-many head benefits more from these changes, achieving improved performance over YOLO-World. However, the one-to-one head still struggled with alignment, particularly in rare class detection.
To address this, we calibrate text encoder with AdaDL, making both image and text encoder learn decision boundaries simultaneously, which attains significant improvements. 

As shown in \cref{tab:ablation-align}, we compare the different metheds for AdaDL to calibrate text encoder. Performing full fine-tuning improves overall accuracy but reduces performance on rare classes, likely due to overfitting. We assume this is caused by the large gap between the number of image and text training parameters. Introducing parameter-efficient methods like prompt tuning~\cite{coop} and deep prompt tuning~\cite{maple} significantly improved alignment, enabling the one-to-one head to match the one-to-many head in performance. And as the training parameters increase, the performance also improves. Finally, using LoRA for text encoder across all projection layers further adapt text information to be region-aware. This approach yielded the best overall results and was adopted for our final experiments.

\begin{figure*}[t]
    \centering
    \includegraphics[width=1\linewidth]{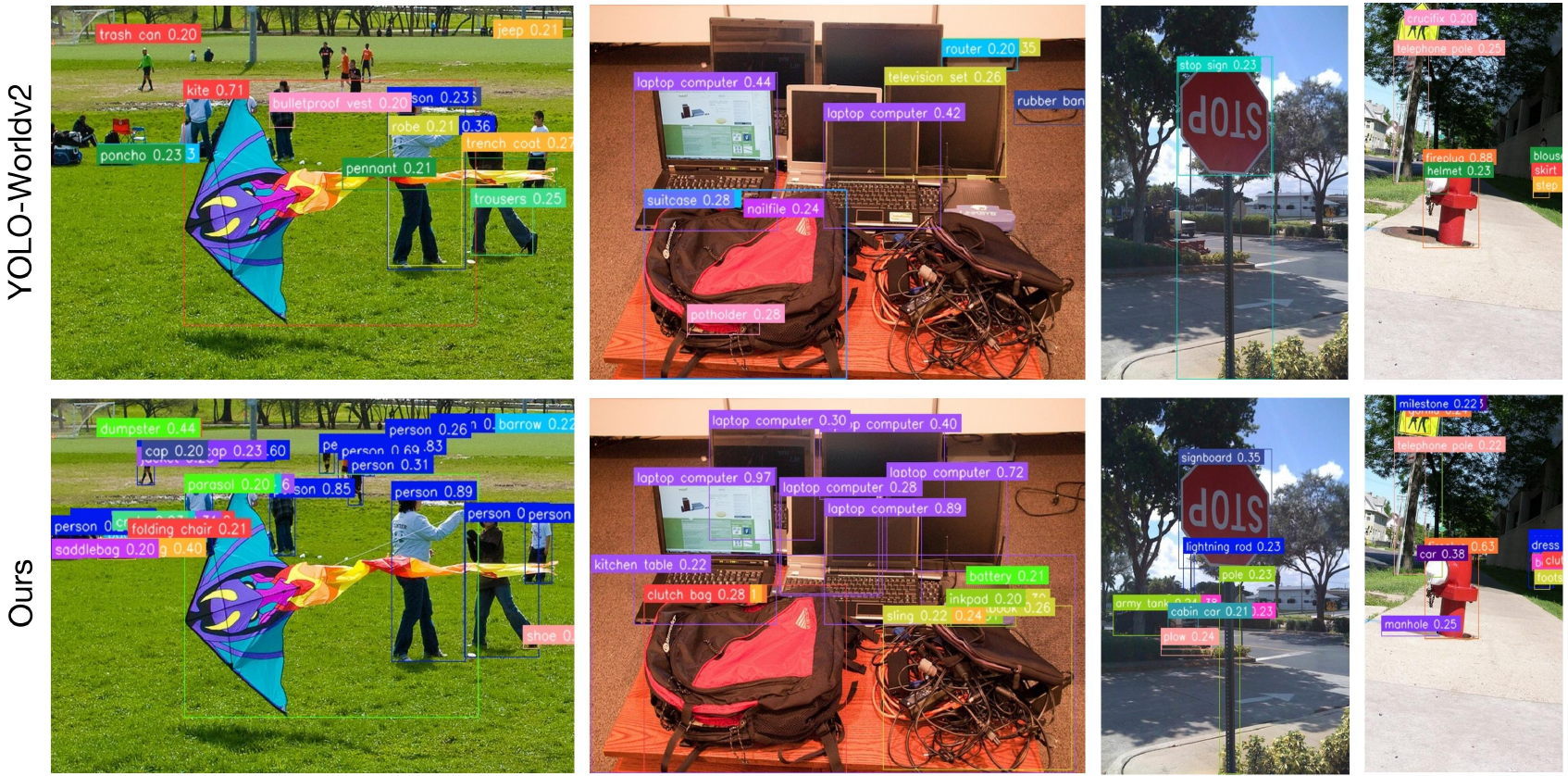}
    \caption{\textbf{Visualization Results on Zero-shot Inference on LVIS.} We present visualization results with YOLO-Worldv2 both in \textit{small} scale, using LVIS 1023 class names as text prompts. The model pretrained with our strategy demonstrates exceptional capability in detecting objects within complex scenes and recognizing a broader range of novel classes.}
    \label{fig:ov}
\end{figure*}

\begin{table}[t]
\caption{\textbf{Ablations on Pre-training Settings.} * means reproduced version in our experiment setting. \textit{w/o VL-PAN} denotes eliminating RepVL-PAN structure in YOLO-Worldv2.}
\label{tab:ablation-pt}
\centering
\resizebox{\columnwidth}{!}{
\begin{tabular}{ccccc} 
\toprule
Method        & $AP$ & $AP_r$       & $AP_c$       & $AP_f$        \\ 
\hline
YOLO-World  & 24.3       & 16.6      & 22.1      & 27.7       \\
YOLO-Worldv2  & 22.7       & 16.3      & 20.8      & 25.5       \\
YOLO-Worldv2* & 23.5       & 16.7      & 20.2      & 27.7       \\
w/o VL-PAN    & 22.8       & 17.1      & 19.4      & 26.8       \\ \hline
+Dual-Head Match       & 21.5/23.3  & 12.4/17.3 & 18.7/20.4 & 25.6/26.9  \\
+AdaDL         & 26.2/27.4  & 24.1/26.0 & 24.9/25.6 & 27.7/29.3  \\
\bottomrule
\end{tabular}
}
\end{table}

\begin{table}[t]
\caption{\textbf{Ablations on AdaDL Methods.} We ablate different methods for adaptive decision learning.}
\label{tab:ablation-align}
\centering
\resizebox{\columnwidth}{!}{
\begin{tabular}{ccccc} 
\toprule
Method            & $AP$ & $AP_r$       & $AP_c$       & $AP_f$        \\ 
\hline
Frozen           & 21.5/23.3  & 12.4/17.3 & 18.7/20.4 & 25.6/26.9  \\
Full Fine-tune    & 23.2/23.8  & 13.7/13.3 & 20.5/20.8 & 27.2/28.4  \\
Prompt  & 24.1/25.1  & 18.5/19.1 & 21.9/23.2 & 27.2/28.0  \\
Deep Prompt & 24.5/25.4  & 19.5/18.5 & 22/23.4   & 27.6/28.5  \\
LoRA    & 26.2/27.4  & 24.1/26.0 & 24.9/25.6 & 27.7/29.3  \\
\bottomrule
\end{tabular}
}
\end{table}

\textbf{Open-World Detection:}
We compared the performance of close-set YOLOv10 trained with unknown class labels (oracle) and zero-shot performance of our open-vocabulary model on the M-OWODB dataset. The results show in \cref{tab:ablation-owod}, even in a zero-shot setting, our open-vocabulary model achieves higher known class accuracy than the oracle-trained YOLOv10 model. Moreover, when we only simply use vanilla ``object" as text input, it achieves better Unknown recall than traditional owod methods, which further validates the effectiveness of our open-vocabualry methods. By applying our wildcard embeddings, the model’s unknown detection capability is fully unlocked, surpassing the performance of models trained with oracle supervision on unknown labels across different tasks. And as the model scales up, its ability to detect known and unknown class increases simultaneously.

\subsection{Qualitative Result}

For the open-vocabulary model, we input the 1,023 category names from the LVIS dataset as prompt, comparing the zero-shot performance on LVIS with YOLO-Worldv2, as shown in \cref{fig:ov}.
It shows that our AdaDL strategy enhances the model's decision boundaries to detect objects of varying sizes, distances, or those partially occluded, with higher confidence scores. Moreover, the improved alignment between visual and calibrated semantic information enables the model to correctly classify detected objects, capturing more diverse categories.

\begin{figure*}[t]
    \centering
    \includegraphics[width=1\linewidth]{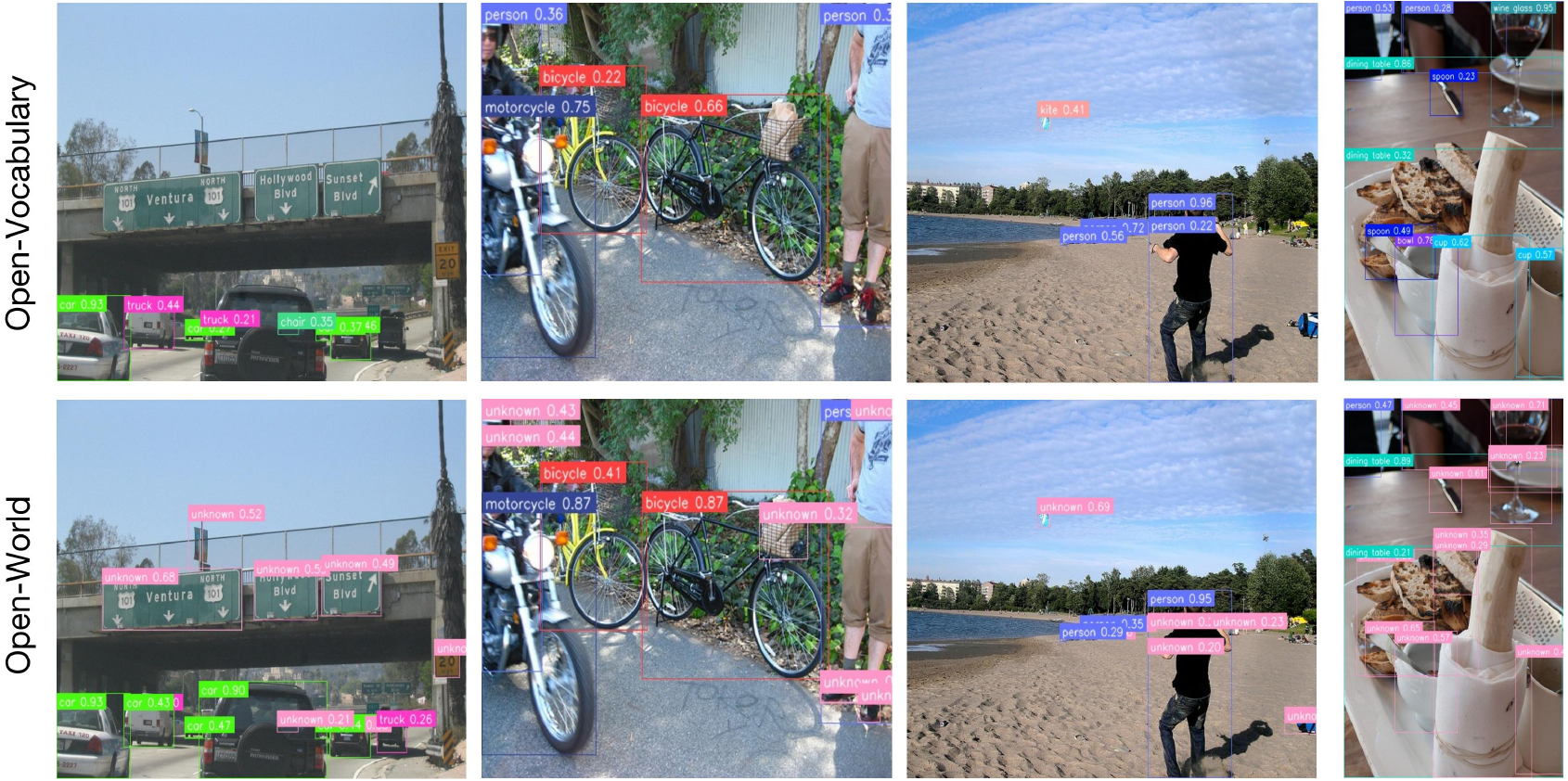}
    \caption{\textbf{Visualization Results on M-OWODB.} Compared to the open-vocabulary model using prompts with all 80 classes, our approach that extends to open-world only employs 40 class embeddings with an additional ``unknown'' wildcard. }
    \label{fig:ow}
\end{figure*}

In \cref{fig:ow}, we compare the performance of an open-vocabulary model using text embeddings for all 80 known classes in the M-OWODB dataset with our model, which uses text embeddings for only half of the known classes (similar to the Task 2 scenario) and an extra ``unknown'' wildcard to detect unknown objects. The results demonstrate that our model not only identifies the remaining 40 unknown classes without corresponding text inputs but also detects additional objects.
This indicates that the ``unknown'' wildcard effectively retains the rich semantic knowledge from pretraining while learning downstream task-specific knowledge, showcasing strong generalization capabilities that align with real-world requirements.

%% file: sec/5_Conclusion.tex
\section{Conclusion}
\label{sec:conlusion}

In this work, we propose Universal Open-World Object Detection (Uni-OWD), a new paradigm to tackle the challenges of dynamic object categories and unknown target recognition within a unified framework. To address this, we introduce YOLO-UniOW, an efficient solution based on the YOLO detector. Our framework incorporates several innovative strategies: the Adaptive Decision Learning (AdaDL) strategy, which seamlessly adapts decision boundaries for Uni-OWD tasks, and Wildcard Learning, using a ``unknown" wildcard embedding to enable the detection of unknown objects, supporting iterative vocabulary expansion without incremental learning.
Extensive experiments across benchmarks for both open-vocabulary and open-world object detection validate the effectiveness of our approach. The results demonstrate that YOLO-UniOW significantly outperforms state-of-the-art methods, offering a versatile and superior solution for open-world object detection. This work highlights the potential of our framework for real-world applications, paving the way for further advancements in this evolving field.